\pdfoutput=1
\documentclass[11pt]{article}

\usepackage[final]{acl}

\usepackage{times}
\usepackage{latexsym}

\usepackage[T1]{fontenc}
\usepackage[utf8]{inputenc}

\usepackage{microtype}

\usepackage{inconsolata}
\usepackage{microtype}

\usepackage{inconsolata}

\usepackage{graphicx}

\usepackage{algorithm}
\usepackage{algorithmic}
\usepackage{color}
\usepackage{mathrsfs}
\usepackage{amsmath}
\usepackage{amsfonts}
\usepackage{multirow} 
\usepackage{subfigure} 
\usepackage{booktabs}

\usepackage{amssymb}
\usepackage{caption}
\usepackage{hyperref}

\usepackage[utf8]{inputenc} % allow utf-8 input
\usepackage[T1]{fontenc}    % use 8-bit T1 fonts
\usepackage{url}            % simple URL typesetting
\usepackage{booktabs}       % professional-quality tables
\usepackage{amsfonts}       % blackboard math symbols
\usepackage{nicefrac}       % compact symbols for 1/2, etc.
\usepackage{microtype}      % microtypography
\usepackage{xcolor}         % colors
\newcommand{\ourbench}{\textsc{IW-Bench}}
\usepackage{enumitem} % For custom enumeration

\usepackage{graphicx}
\usepackage{subfigure}
\usepackage{booktabs} % for professional tables
\usepackage{amsmath}
\usepackage{natbib}
\usepackage{tcolorbox}

\usepackage{graphicx}

\newtcolorbox{mybox}[1][]{
  arc=1mm,
  boxrule=1pt,
  colback=gray!20, % 更改背景颜色为灰色
  colframe=black!80,
  fonttitle=\bfseries,
  fontupper=\small, % 更改字体大小为小号
  title={Content},
  left=1mm,
  right=1mm,
  top=1mm,
  bottom=1mm
}

\title{IW-Bench: Evaluating Large Multimodal Models for Converting Image-to-Web}

\author{
 \textbf{Hongcheng Guo}\textsuperscript{1},
 \textbf{Wei Zhang}\textsuperscript{1},
 \textbf{Junhao Chen}\textsuperscript{3}, 
 \textbf{Yaonan Gu}\textsuperscript{5},
 \textbf{Jian Yang}\textsuperscript{2}, 
 \textbf{Junjia Du}\textsuperscript{4}, \\
  \textbf{Shaosheng Cao},
    \textbf{Binyuan Hui}\textsuperscript{2},
       \textbf{Tianyu Liu}\textsuperscript{2},
         \textbf{Jianxin Ma}\textsuperscript{2},
         \textbf{Chang Zhou}\textsuperscript{2},
  \textbf{Zhoujun Li}\textsuperscript{1}\thanks{Corresponding author.},
  \\
 \\
 \textsuperscript{1}Beihang University
 \textsuperscript{2}Alibaba Group
 \textsuperscript{3}Tsinghua University \\ 
 \textsuperscript{4}Nanyang Technological University
 \textsuperscript{5}National University of Singapore\\
  \texttt{\{lizj, hongchengguo\}@buaa.edu.cn},
}

\begin{document}
\maketitle
\begin{abstract}
Recently advancements in large multimodal models have led to significant strides in image comprehension capabilities. Despite these advancements, there is a lack of the robust benchmark specifically for assessing the Image-to-Web conversion proficiency of these large models. Primarily, it is essential to ensure the integrity of the web elements generated. These elements comprise visible and invisible categories. Previous evaluation methods (e.g., BLEU) are notably susceptible to significant alterations due to the presence of invisible elements in Web. Furthermore, it is crucial to measure the layout information of web pages, referring to the positional relationships between elements, which is overlooked by previous work. To address challenges, we have curated and aligned a benchmark of images and corresponding web codes (\textbf{\ourbench{}}). Specifically, we propose the \textbf{Element Accuracy}, which tests the completeness of the elements by parsing the Document Object Model (DOM) tree. \textbf{Layout Accuracy} is also proposed to analyze the positional relationships of elements by converting DOM tree into a common subsequence. Besides, we design a five-hop multimodal Chain-of-Thought Prompting for better performance, which contains five hop: 1) SoM prompt injection. 2) Inferring Elements. 3) Inferring Layout. 4) Inferring Web code. 5) Reflection.
Our benchmark comprises 1200 pairs of images and web codes with varying levels of difficulty. We have conducted extensive experiments on existing large multimodal models, offering insights into their performance and areas for improvement in image-to-web domain.~\footnote{We provide code and dataset at \url{https://github.com/HC-Guo/IWBench}
} 

\end{abstract}

\section{Introduction}

\begin{figure}[t]
    \centering {\includegraphics[width=\linewidth]{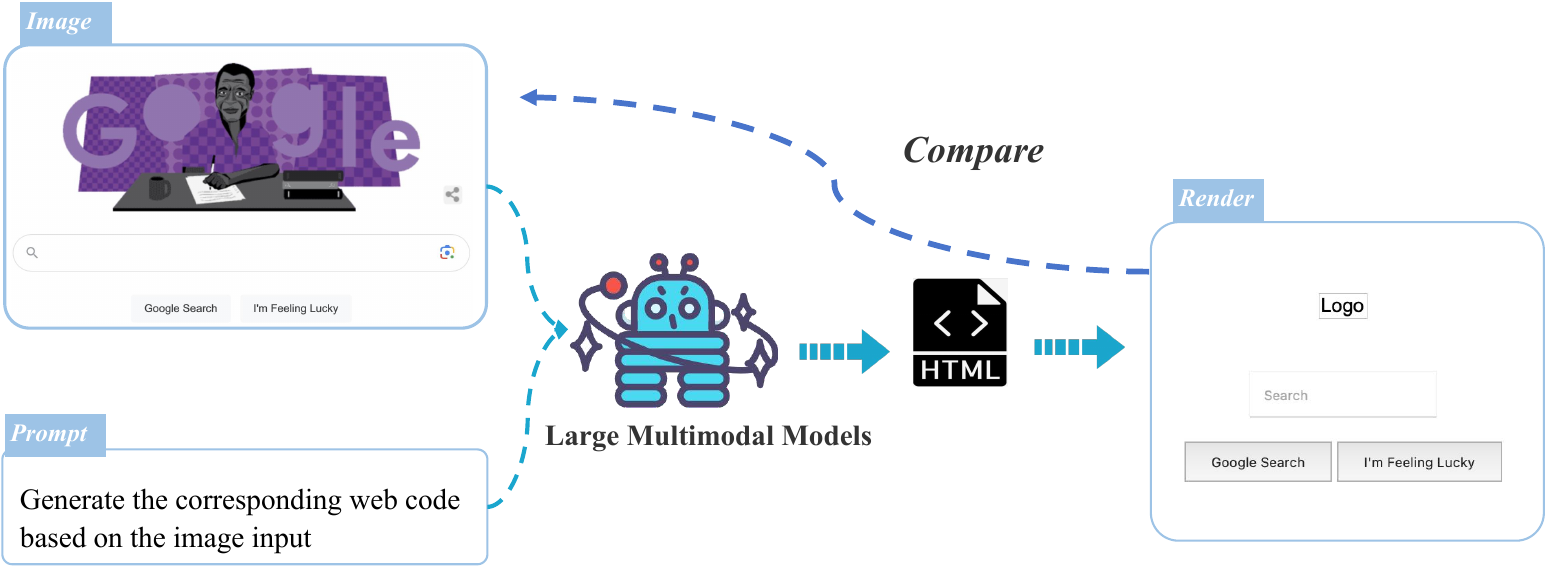}}
    \caption{The process of Image-to-Web. We prompt a large multimodal model to generate web code based on the input image. Finally, we need to compare whether the newly rendered result is consistent with the input image.}
    \label{intro}
\end{figure}

The development of large multimodal models has emerged as a new trend, starting with GPT-4~\cite{openai2023gpt4v}. An increasing number of multimodal models have been introduced~\cite{gemini,liu2023llava,zhu2023minigpt,qwenvl,instructblip}, extending the powerful comprehension capabilities of large language models to multiple tasks~\cite{lu2018rvqa,som,sidorov2020textcaps}. Recently, the task of converting images into web code~\cite{websight,imagecode1,pix2code} has garnered significant attention due to its impressive performance. In Figure~\ref{intro}, this task tests the synthesis abilities of large multimodal models, encompassing the fine-grained recognition of components within images, the assessment of the relative positioning of elements in webpages, and the capability to generate code. The prowess of large models in generating front-end code has been a source of astonishment, yet, notably, there has been almost no evaluation benchmark related to large multimodal models for this domain.

Constructing a benchmark for appraising Image-to-Web tasks is markedly more intricate than the conventional programming challenges~\cite{imagecode7}. This heightens complexity stems from the multifaceted nature of web code, encompassing HTML, CSS, and JavaScript. It demands scrutiny not only of functional correctness but also of visual elements like layout, design, and user interaction. Predominant evaluation frameworks for the Image-to-Code task~\cite{imagecode1,imagecode3} often hinge on similarity metrics such as BLEU. Nevertheless, these metrics encounter substantive limitations in the context of web evaluation. The intricate web of JavaScript functionalities and CSS styling options means that disparate web codes can produce indistinguishable visual outcomes. Consequently, such conventional metrics falter in precisely assessing the completeness of elements depicted in images and their layout intricacies. A further complication arises from non-visible elements, introducing a fragility in these evaluation methods. Varied development ways can lead to substantial discrepancies in these non-visible components, even among web pages that appear visually identical. Recent WebSight~\cite{websight} is finetuned on numerous image-code pairs, however it has not conducted performance evaluations. These challenges underscore the imperative for a more sophisticated and comprehensive benchmark.

To address the challenges, we introduce \ourbench{} to assess the capabilities of large multimodal models in the Image-to-Web. Firstly, we design a data construction pipeline, comprising 1200 entries of three difficulty levels: Level I (\textit{simple}), Level II (\textit{medium}), and Level III (\textit{complex}). For metric innovation, we propose \textbf{Element Accuracy} and \textbf{Layout Accuracy}. Specifically, to effectively measure the completeness of web elements, the element accuracy tests the completeness of the elements from six dimensions (\texttt{Tag, Text Content, Attribute, Style, JavaScript, Children}) by parsing the Document Object Model (DOM) tree. To analyze the relative positional relationships of elements, we first traverse the DOM Tree in the same manner (in-order traversal) to obtain a list of elements, then we calculate the overlap between the common sub-sequence and the ground truth for the layout accuracy. Moreover, we establish a effective five-hop multimodal chain-of-thought method to enhance the model performance, including five specialized hops: SoM prompt injection, Inferring Elements, Inferring Layout, Inferring Web code, and Reflection.

To summarize:
\begin{itemize}
    \item \textbf{Benchmark Construction.} We have meticulously curated 1200 three challenging levels of image-web pairs for our benchmark dataset. This dataset serves as a rigorous benchmark for assessing the capabilities of large multimodal models in the task of converting images to web code.
    \item \textbf{Metric Innovation.} We introduce innovative metrics to evaluate web elements and layout information accurately. Specifically, we have developed the Element Accuracy metric to assess the fidelity of web elements and the Layout Accuracy metric to evaluate the precision of layout information.
    \item \textbf{Multimodal Chain-of-Thought.} The five-hop multimodal Chain-of-Thought method is proposed by us, significantly enhancing the performance in image-to-web domain.
    \item \textbf{Extensive Evaluation.} A substantial evaluation of large multimodal models has been conducted, showcasing their capabilities and limitations in our context. Besides, ablation on our five-hop chain-of-thought demonstrates the effect of our method.
\end{itemize}

\section{Related work}

Multimodal benchmarks for large multimodal models~\citep{bitton2023visit,yu2023mm,liu2023mmbench,xu2023lvlm,shao2023tiny} have assess the instruction-following and reasoning capabilities. As these foundation models become more relevant to real-world applications, unlike prior work, we propose \ourbench{} to benchmark their generation capabilities of the hot Image-to-Web area on a diverse set of visual contexts. 
To the best of our knowledge, WebSight~\cite{websight} is the only relevant benchmark. However, it is more like a fine-tuning dataset for large multimodal models rather than a benchmark for evaluation. The detailed comparison is in Table~\ref{table:compare}. More related work on image-to-web, large multimodal models, and chain-of-thought methods are in Appendix~\ref{appendix:relatedwork}.

\begin{table}[htp]
\centering
\resizebox{1\linewidth}{!}{
\begin{tabular}{ccc}
\toprule
           Feature & IW-Bench & Websight \\
\midrule
    Language Coverage & Chinese and English & English \\ \midrule
    Real Web Page Samples & \color{green}$\checkmark$ & \color{red}$\times$ \\ \midrule
    Data De-identification & \color{green}$\checkmark$ & \color{red}$\times$ \\ \midrule
    Enhanced Data Quality & Automated and Manual Review & Automated Review \\ \midrule
    Quantitative Analysis & \color{green}$\checkmark$ & \color{red}$\times$ \\ \midrule
    Human Evaluation & \color{green}$\checkmark$ & \color{red}$\times$ \\ \midrule
    Evaluation Metrics & Element and Layout Accuracy & \color{red}$\times$ \\

\bottomrule
\end{tabular}}
\caption{Comparison between \ourbench{} and Websight.}
\label{table:compare}
\end{table}

\section{\ourbench{}}

\begin{figure*}[htbp]
    \centering {\includegraphics[width=0.85\textwidth]{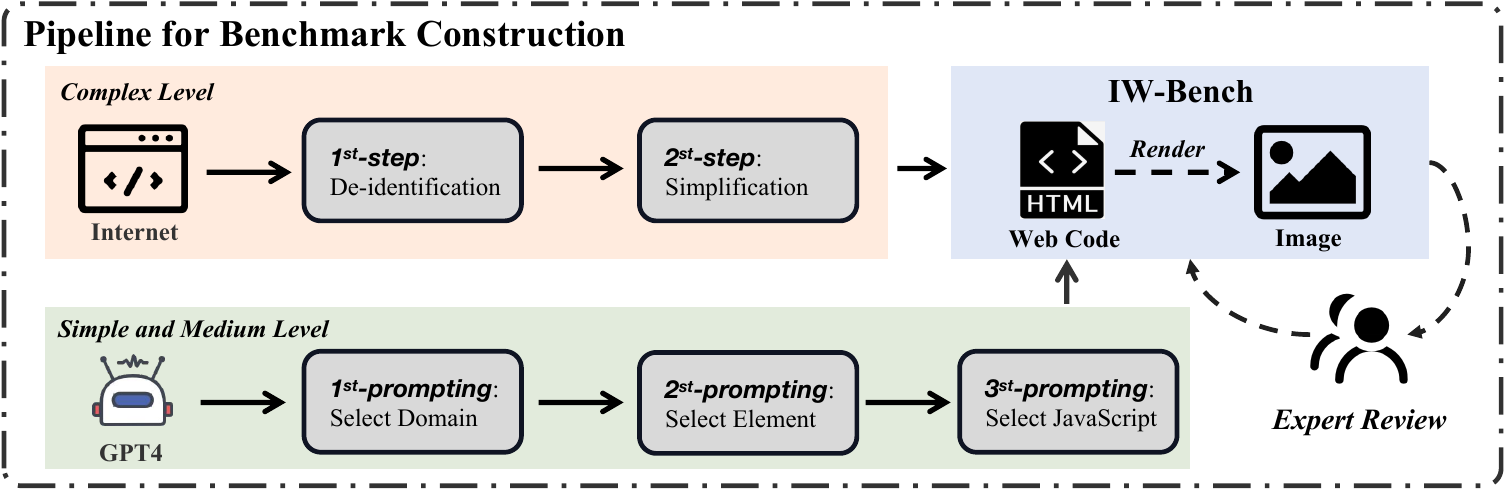}}

    \caption{Benchmark Construction. This pipline illustrates the multi-step process used to construct \ourbench{} for web code and images of varying complexity levels.}
    \label{benchmark_construction}
\end{figure*}

\paragraph{Overview:} We design a pipeline for the construction of \ourbench{} in Figure~\ref{benchmark_construction}. Specifically, to obtain complex-level data, we crawl publicly available web pages and perform de-identification and simplification. For simple and medium level data, we prompt GPT-4~\cite{openai2023gpt4} to generate them. Finally, we filter out some low-quality samples from the obtained image-code pairs. Quantity of \ourbench{} is in Table~\ref{bench:number} and word cloud is in Figure~\ref{fig:wordcloud}. We present examples in Appendix~\ref{case:all}.

% Thus our data collection process involves the acquisition of test data from different subdomains: photography, geography, company, shopping, blog and so on. 

\begin{figure}[t]
    \centering {\includegraphics[width=1\linewidth]{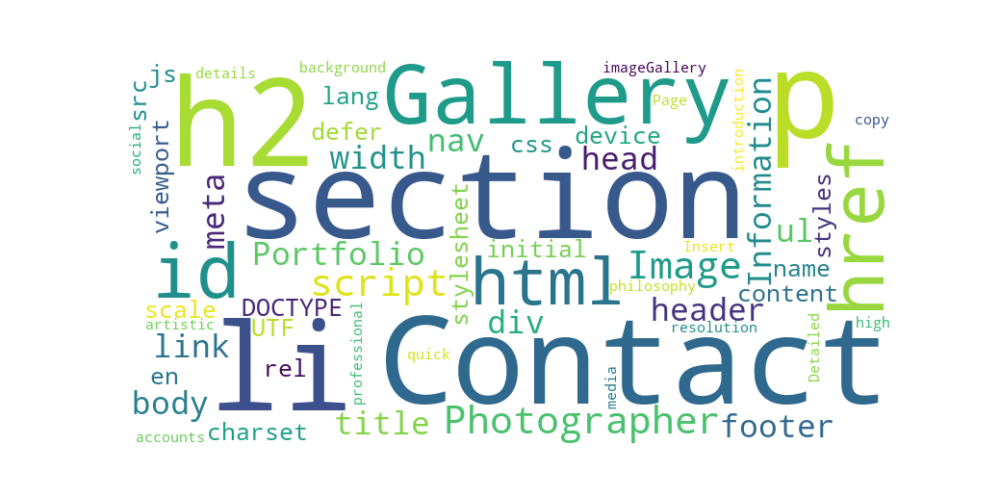}}
    \caption{Wordcloud. The key words in \ourbench{} are related to the web and internet, such as 'html', 'header'.}
    \label{fig:wordcloud}
\end{figure}

\begin{table}[htp]
\centering
\resizebox{0.9\linewidth}{!}{
\begin{tabular}{lcccccc}
\toprule
          &  & \#Level I (Simple) &\#Level II (Medium) &\#Level III (Complex) \\
\midrule

        & Number  & 340  & 645 & 215  \\
\midrule
        & Source  & GPT4  & GPT4 & Internet  \\

\bottomrule
\end{tabular}}
\caption{Quantity of \ourbench{}. Varied levels of complexity exist within dataset with varying numbers. The sources of \ourbench{} are GPT4 and real web pages.}
\label{bench:number}
\end{table}

% onsisting of xxx entries, medium-level part consisting of xxx pairs, and complex-level part consisting of xxx entries 

\subsection{Data Collection}

We initially intend to assign the task of determining complexity to human workers and exclusively use real web. However, tests reveal that websites in real-world scenarios are too complex, resulting in uniformly poor evaluation with minimal differentiation between models. The criteria for different complexity levels is in Table~\ref{tab:webpage_complexity}.

\begin{table}[h!]
    \centering
    \resizebox{1\linewidth}{!}{\begin{tabular}{cccc}
        \toprule
        \textbf{Web Page Complexity Level} & \textbf{Number of Elements} & \textbf{Number of Attributes} & \textbf{JavaScript Usage} \\
        \midrule
        Simple & 1-20 & 1-20 & None or minimal \\
        \midrule
        Medium & 20-60 & 20-60 & Moderate \\
        \midrule
        Complex & 61+ & 61+ & High\\
        \bottomrule
    \end{tabular}}
\caption{Web Page Complexity Levels and Their Characteristics}
\label{tab:webpage_complexity}
\end{table}

To better quantify differences among models, we opt to create simpler data by prompting GPT4~\cite{openai2023gpt4} following the previous work~\cite{self_instruct,xu2023baize}. The process begins with GPT-4 randomly selecting a relevant domain in Appendix~\ref{prompt:promptdomain}. Based on the desired complexity, GPT-4 then chooses specific elements. Finally, it selects the JavaScript events in Apppendix~\ref{promt:java} to be integrated into the web page. In \ourbench{}, 18\% of the data comes from real web pages, covering a wide range of types, including public repositories, open-source projects, and websites representing various content categories. The data collection process strictly follows the robots.txt protocol and complies with website policies. To extract images from web code, we use Playwright~\footnote{https://github.com/microsoft/playwright}.

% \begin{table}[htp]
% \centering
% \caption{Number of \ourbench{}. Varied levels of complexity exist within dataset with varying numbers.}
% \resizebox{0.95\linewidth}{!}{
% \begin{tabular}{lcccccc}
% \toprule
%           & \#Total number & \#Level I (Simple) &\#Level II (Medium) &\#Level III (Complex) \\
% \midrule

%         & 1000  & 140  & 645 & 215  \\

% \bottomrule
% \end{tabular}}
% \label{bench:number}
% \end{table}

\subsection{Data Processing}
The processing of real web page data includes two parts: de-identification and simplification.
For de-identification, we replace the image components and modify the logos within our dataset to avoid copyright or other legal concerns. More details are in Appendix~\ref{app:deid}. 
For simplification, this involves the removal of non-essential invisible elements such as whitespace, code comments, and styles. We also streamline HTML by condensing verbose code and transferring inline styles to external stylesheets, which contributes to a more uniform and computationally efficient dataset. Further, we standardize visible elements like images and text by replacing them with placeholders and normalizing stylistic features. 

We carefully validate the simplifications to ensure that the functionality and layout of the original webpage remain unaffected. Specifically, we start by launching a headless browser using Playwright and capturing the initial screenshot of the webpage. After that, we employ BeautifulSoup to parse the HTML document. Subsequently, we traverse through all HTML elements, with the exception of fundamental tags such as html, head, body, title, and meta, as these are typically crucial components of the structure. We initiate the removal process and then scrutinize whether there is any discernible alteration in the visual presentation. To determine this, we render the modified HTML content and compare it with the original one. If the new screenshot matches the original one, we retain the deleted element and document it accordingly. The time cost of simplification is in Table~\ref{Simplification Time Cost}.

\begin{table}[htp]
\centering
\resizebox{0.8\linewidth}{!}{
\begin{tabular}{cc}
\toprule
           Web Page Complexity & Average Simplification Time (seconds) \\
\midrule

     Simple  & 44.3\\
    \midrule
     Medium  & 131.5 \\
    \midrule
     Complex & 238.2  \\

\bottomrule
\end{tabular}}
\caption{Simplification Time Cost.}
\label{Simplification Time Cost}
\end{table}

\subsection{Expert Review}

Establishing \ourbench{} involves substantial manpower. In the collection phase, 20 web engineers assess complexity based on HTML tag variety, DOM tree depth, and scripting presence. Each data point is cross-validated by three engineers, with disagreements resolved through training, detailed guidelines, and a majority-rule consensus mechanism. Low-quality examples are re-evaluated until agreement is achieved.

\subsection{Metric Design}

To evaluate the fidelity of web elements, we propose two novel metrics: \textbf{Element Accuracy (EA)} and \textbf{Layout Accuracy (LA)}, which are designed to compare visible and invisible elements in web pages. The evaluation process begins with constructing a \textbf{Document Object Model (DOM)} tree for the test and reference web pages, followed by a structured traversal and analysis of the elements. The DOM tree traversal systematically inspects each element, collecting detailed information such as tag types, attributes, styles, JavaScript bindings, and hierarchical relationships. This unified approach enables the evaluation of visible elements (e.g., text, images, buttons) alongside invisible ones (e.g., \texttt{<script>} tags, metadata).

\subsubsection{Element Accuracy (EA)}
Element Accuracy measures the similarity between corresponding elements in the test and reference web pages based on six key attributes: \textbf{Tag}, \textbf{Text Content}, \textbf{Attributes}, \textbf{Style}, \textbf{JavaScript}, and \textbf{Children}. Let the test element set be \( E_{\text{test}} \), and the label element set be \( E_{\text{label}} \). For each test element \( E_j \), an average score \( S_j \) is computed across these six perspectives. The Element Accuracy is defined as:

\begin{equation}
EA = \frac{\sum_{j \in E_{\text{test}}} \mathbb{1}(E_j > T)}{|E_{\text{label}}|}
\end{equation}
where \( T \) is the threshold, and \( \mathbb{1}(E_j > T) \) is an indicator function that equals 1 if \( E_j \) exceeds \( T \), and 0 otherwise.

The similarity scores for each perspective are computed as follows:
\begin{itemize}
    \item \textbf{Tag}: Tags are compared by their names: 
    \begin{equation}
    S_{\text{tag}} = 
    \begin{cases} 
    1 & \text{if } \text{tag}_{\text{test}} = \text{tag}_{\text{label}}, \\
    0 & \text{otherwise}.
    \end{cases}
    \end{equation}
    
    \item \textbf{Text Content}: The similarity of text content between two elements is computed by SequenceMatcher (SM)~\footnote{https://github.com/m-matelski/mdiff}:
    \begin{equation}
    S_{\text{text}} = \text{SM}(\text{content}_{\text{test}}, \text{content}_{\text{label}})
    \end{equation}

    \item \textbf{Attributes}: A predefined mapping is used for each tag type. If the corresponding attributes of two elements match exactly, they receive a higher score:
    \begin{equation}
    S_{\text{attr}} = \frac{\sum_{k \in A} \mathbb{1}(\text{attr}_k^{\text{test}} = \text{attr}_k^{\text{label}})}{|A|}
    \end{equation}
    where \( A \) is the set of relevant attributes for the tag.

    \item \textbf{Style}: Key style properties (e.g., color, font-size) are filtered out, ignoring default values. Matching values contribute to the score:
    \begin{equation}
    S_{\text{style}} = \frac{\sum_{p \in P} \mathbb{1}(\text{style}_p^{\text{test}} = \text{style}_p^{\text{label}})}{|P|}
    \end{equation}
    where \( P \) is the set of key style properties.

    \item \textbf{JavaScript}: Event bindings on two elements are compared (e.g., \texttt{onclick}, \texttt{onload}). Let \( J \) be the set of relevant events. The JavaScript score is:
    \begin{equation}
    S_{\text{js}} = \frac{\sum_{e \in J} \mathbb{1}(\text{event}_e^{\text{test}} = \text{event}_e^{\text{label}})}{|J|}
    \end{equation}

    \item \textbf{Children}: The hierarchical structure is evaluated using a tree-edit distance algorithm.
\end{itemize}

The average score for each test element \( E_j \) is:
\begin{equation}
E_j = \frac{\sum_{i=1}^{n} S_i}{n},
\end{equation}
where \( n = 6 \) represents the number of evaluated attributes: \textit{Tag, Text Content, Attribute, Style, JavaScript, Children}.

\subsubsection{Layout Accuracy (LA)}
Layout Accuracy quantifies the structural similarity between the layouts of web pages by applying the \textbf{Longest Common Subsequence (LCS)} on their element lists. The layout is linearized into a sequence of elements for comparison.

Let \( L_1 \) be the list of web elements from the label HTML, and \( L_2 \) be the list from the generated Web code. The layout accuracy is defined as:
\begin{equation}
LA = \frac{\text{LCS}(L_1, L_2)}{\text{Len}(L_1)}
\end{equation}

\begin{equation}
\small
\text{LCS}(L_1, L_2) = \max \left( \delta(e_1, e_2) + \text{LCS}(L_1', L_2') \right)
\end{equation}

where \( \text{LCS}(L_1, L_2) \) is the length of the longest common subsequence of elements between \( L_1 \) and \( L_2 \), \( \text{Len}(L_1) \) is the total length of \( L_1 \). \( \delta(e_1, e_2) \) be a similarity function between two elements \( e_1 \) and \( e_2 \). \( L_1' \) and \( L_2' \) are the remaining subsequences after matching \( e_1 \) and \( e_2 \).

Further details about style properties are in Appendix~\ref{app:style}, and JavaScript events are in Appendix~\ref{app:jsevent}.

\begin{figure*}[htbp]
    \centering {\includegraphics[width=0.85\textwidth]{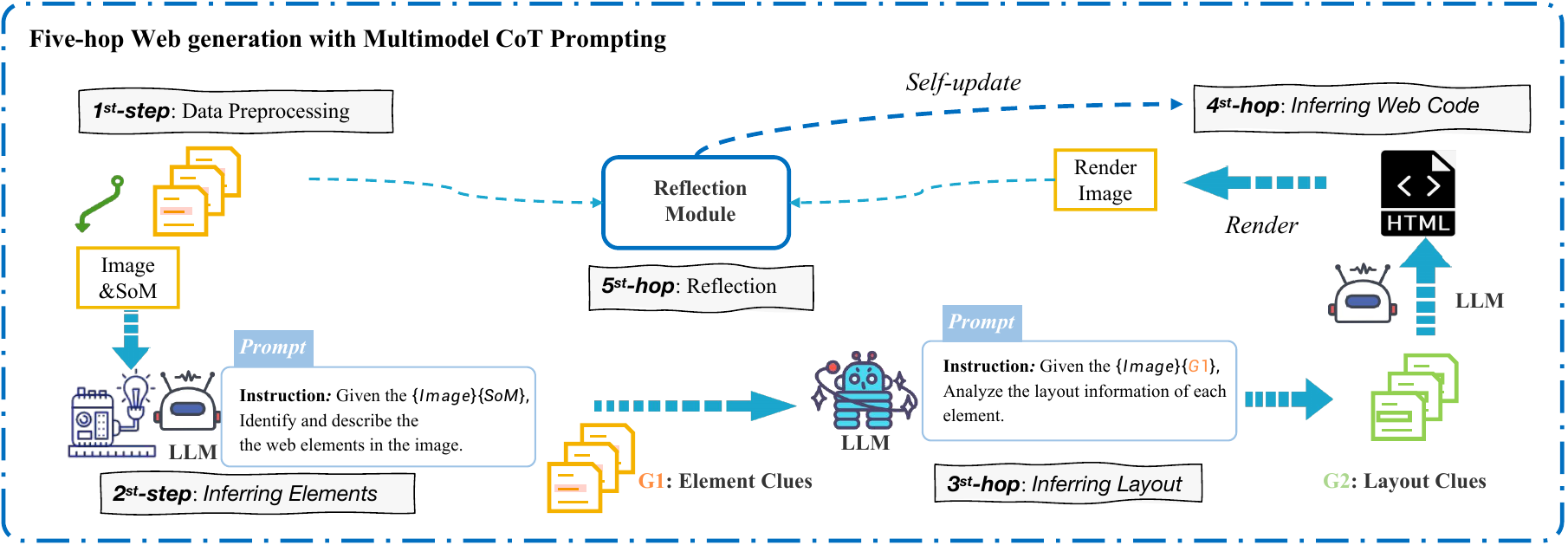}}

    \caption{Overview of Five-hop Multimodal Chain-of-Thought Prompting. Our method contains five hop: 1) SoM prompt injection. 2) Inferring Elements. 3) Inferring Layout. 4)Inferring Web code. 5) Reflection.}
    \label{framework}
\end{figure*}

\begin{figure}[htbp]
    \centering {\includegraphics[width=1\linewidth]{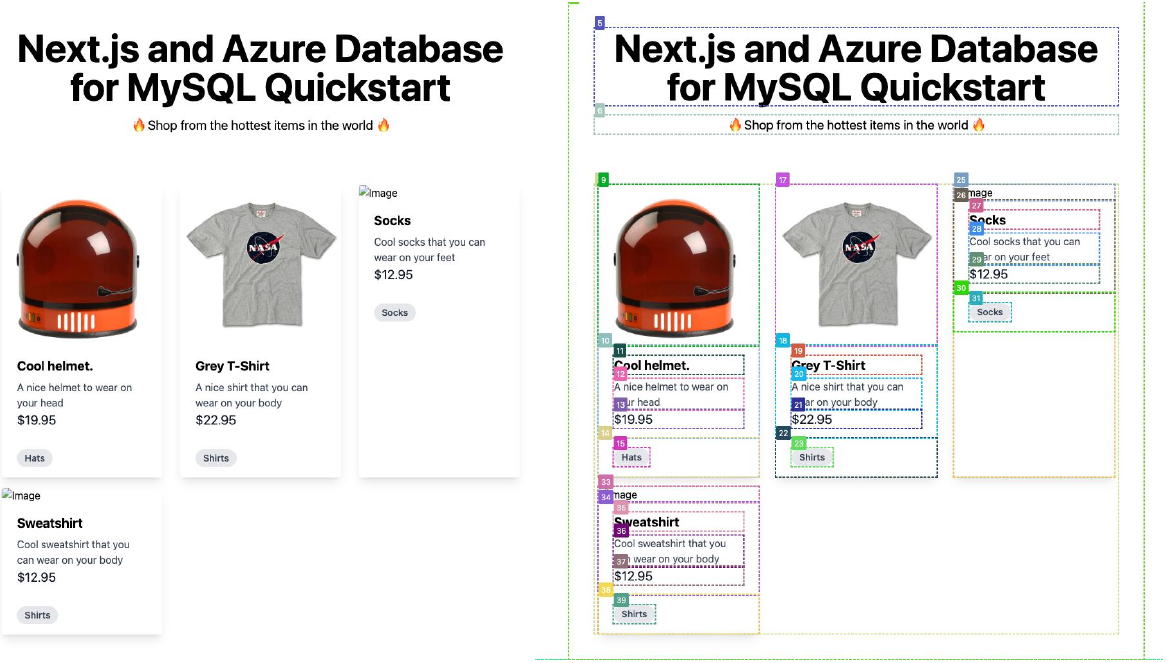}}

    \caption{Example of SOM prompt injection. The image on the left is the original web page, and the image on the right is the rendered web page after injection.}
    \label{fig:som}
\end{figure}

\section{Five-hop multimodal Chain-of-Thought}
In this section, we design a five-hop multimodal Chain-of-Thought Prompting for image-to-html task in Figure~\ref{framework}, which contains five hop: 1) SoM prompt injection. 2) Inferring Elements. 3) Inferring Layout. 4)Inferring Web code. 5) Reflection.

\subsection{SoM Prompt Injection} 
Inspired by prior work~\cite{som}, we enhance images with Scenes of Meaning (SoM) cues, as shown in Figure~\ref{fig:som}. By identifying key elements and integrating cues like text labels, arrows, and highlights, we guide multimodal models to focus on crucial details, improving comprehension and precision.

\subsection{Inferring Elements}

Integrating SoM prompts enhances element inference in images, improving categorization by type and functional role. This aids in understanding both visible and invisible elements and their contributions to the overall context.

\begin{mybox}
\textbf{Infer Element Prompt}: First, analyze this screenshot of the webpage, please try your best to identify and describe this webpage’s functions and its web elements. Some of these elements have been numerically labeled in sequence with bounding boxes.
\label{infer_element}
% \vspace{-3mm}
\end{mybox}

\subsection{Inferring Layout}
In the third hop, models infer layout information, analyzing spatial arrangement, alignment, proximity, and hierarchy of web elements. This step uncovers functional relationships and visual impact, helping models understand UI structure, user interaction flow, and design principles to refine content generation.

\begin{mybox}
\textbf{Inferring Layout Prompt}: The second step is to demonstrate the positional relationships of the marked web page elements based on the provided bounding boxes, including the overall layout and the relative positions between elements.
\label{infer_layout}
% \vspace{-3mm}
\end{mybox}

\subsection{Inferring Web Code}

Once we have a detailed understanding of the interface, we can then generate HTML code based on the provided instructions. The HTML code will reflect the layout, style, and interactive features of the interface, ensuring that the final web page visually and functionally aligns with the analyzed design.

\begin{mybox}
\textbf{Inferring Web Code Prompt}: Please as per the above descriptions of the webpage’s overall layout and web elements together with their relative positioning, generate web code for the corresponding original web image by skipping the step of assigning bounding boxes to elements.
\label{infer_Webcode}
\end{mybox}

\subsection{Reflection}
In reflection, we first re-render the generated HTML code and produce screenshots. Then, we require the large multimodal models to compare the element completeness and layout information of two screenshots from an image perspective. Following this comparison, we generate analysis content based on the results and feed this content back into the previous hop, assisting in the generation of code once again. This is an iterative process, and we control the number of iterations with a hyperparameter N. This method ensures that the generated HTML code is not only accurate in terms of code but also visually consistent with the original design, thereby enhancing the quality of the final result. 

\begin{mybox}
\textbf{Reflection and Self-update Prompt}: Please compare the two screenshots of webpages. The latter is the screenshot of the webpage by the web code you just provided. Based on the above web element descriptions and layout information, please identify whether there are missing elements and access whether the layout and elements’ relative positioning are correct. Afterwards, please improve the web code accordingly.
\label{infer_Webcode}
\end{mybox}

\section{Evaluation}

\subsection{Experiment setting}

We select recent multimodal language models as baselines. GPT4V (gpt-4-1106-preview)~\citep{openai2023gpt4v}, Qwen-VL-Chat,Qwen-VL-Plus, Qwen-VL-Max~\cite{qwenvl}, miniGPT-4-LLaMA-2-7B~\cite{zhu2023minigpt}, LLaVA-LLaMA-2-13B~\cite{liu2023llava}, mPLUG-OWL2 \cite{ye2023mplug}, LLaMA-Adapter-V2-7B~\cite{gao2023llamaadapterv2}, WebSight~\cite{websight}, Gemini Pro~\cite{gemini}, Claude3 Opus~\footnote{https://www.anthropic.com/news/claude-3-family}. All experiments are conducted with 16 NVIDIA A100 GPUs (80G). The $T$ in our experiment is 0.9. $N$ is default set to be 3 and we choose the best result. We do not conduct Five-hop MCoT on Websight model as it fails to accommodate simultaneous input of images and text.
% We evaluate the models on \dataset under three setups: (a) \textit{Text-Only LLMs} including ChatGPT \citep{openai2022chatgpt}, GPT-4 \citep{openai2023gpt4}, and Claude-2 \citep{claude2} in zero-shot and two-shot settings with Chain-of-Thought (CoT) \citep{wei2022chain} and Program-of-Thought (PoT) \citep{chen2022program}, (b) \textit{Augmented-LLMs} where the LLMs are provided with additional visual information including the generated image captions from Multimodal Bard \citep{google2023bard} and the detected OCR text from EasyOCR \citep{jaidedai2020easyocr}, (c) \textit{LMMs} that include open-source models such as IDEFICS-9B \citep{laurencon2023obelics}, mPLUG-OWL-LLaMA-7B \citep{ye2023mplug}, miniGPT-4-LLaMA-2-7B \citep{zhu2023minigpt}, LLaMA-Adapter-V2-7B \citep{gao2023llamaadapterv2}, InstructBLIP-Vicuna-7B \citep{instructblip}, LLaVA-LLaMA-2-13B \citep{liu2023llava}, LLaVAR \cite{zhang2023llavar}, and \new{proprietary models such as Bard and GPT-4V. Since GPT-4V does not offer API access, we resorted to manually evaluating it using the playground chatbot}. We provide the prompts for LLMs and the hyperparameters used for LMMs in \S \ref{app:setup}.

\subsection{Experimental Results}
Our results in Table~\ref{maincomparison} are divided into two sections: (1) web code directly generated by the model, and (2) results using the five-hop multi-modal chain-of-thought. We evaluate model in identifying elements and arranging layouts across three complexity levels: simple, medium, and complex.

\subsubsection{Overall Performance Comparison}

WebSight stands out with the highest averages in both element accuracy at 48.9\% and layout accuracy at 47.9\%, which demonstrates the effect of the supervised finetuning. GP4V with Five-hop MCoT also shows significant improvement, with average results of 45.8\% for element accuracy and 44.5\% for layout accuracy, compared to the 30.4\% and 29.4\% without enhancement respectively. The model with the lowest overall performance is LLaMA-Adapter-V2-7B. Models with the same size indeed show varying performances, but an examination of a series of models, such as the Qwen family, reveals that the size of the models significantly affects their effectiveness. This unequivocally proves the effectiveness of our benchmark.

\begin{table*}[t]
\centering
\resizebox{1\textwidth}{!}{
\begin{tabular}{ccccccccc}
\toprule
           Model &\multicolumn{2}{c}{Simple} &\multicolumn{2}{c}{Medium} &\multicolumn{2}{c}{Complex} &\multicolumn{2}{c}{Average} \\

           \cmidrule(lr){2-3}\cmidrule(lr){4-5}\cmidrule(lr){6-7}\cmidrule(lr){8-9}
           &Element Accuracy &Layout Accuracy &Element Accuracy &Layout Accuracy &Element Accuracy &Layout Accuracy &Element Accuracy &Layout Accuracy \\
           
\midrule
    \multicolumn{5}{r}{\hfill \textit{Large Multimodal Models (LMMs)}}\\
    \midrule

   GP4V &46.3 &44.6 &27.4 &26.6 &14.5 & 13.7 &30.4&29.4  \\
                \midrule
    Qwen-VL-Chat&43.8&42.4&24.1&23.3&11.9&11.2&27.4&26.5  \\
    \midrule
    Qwen-VL-Plus&44.4&43.6&25.2&24.2&13.8&13.1&28.6&27.7 \\
    \midrule
    Qwen-VL-Max&59.8&58.8&36.9&35.3&17.6&16.4&39.9&38.5\\
                 
                \midrule
    LLaVA-LLaMA-2-13B&32.1&31.8&18.9&18.1&6.3&5.9&20.3&19.7 \\
                 \midrule
    miniGPT-4-LLaMA-2-7B&20.5 &19.4&13.6&12.8&4.2&4.0&13.8& 13.1  \\
                 
                \midrule
    mPLUG-OWL2 &24.7&23.8&14.6&13.2&7.3&6.0&16.1&14.9    \\
                 
                \midrule
    LLaMA-Adapter-V2-7B&8.6&7.8&4.3&4.4&4.1&4.0&5.5&5.3   \\
                
                \midrule
    Claude 3 Opus &51.5&49.6&30.1&28.4&14.1&13.8&33.3&31.8   \\
                
                \midrule
                
    Gemini Pro  &56.5&54.9&34.1&33.6&15.3&14.5&37.1&36.2  \\
                
                \midrule

WebSight     &64.7&64.3&50.2&49.0&20.4&19.1&\textbf{48.9}&\textbf{47.9}  \\
                 
                \midrule

\multicolumn{6}{r}{\hfill \textit{Large Multimodal Models With Five-hop MCoT (LMMs-CoT) }}\\
    \midrule

    GP4V &69.4&66.4&43.1&42.6&16.7&15.9&45.8&44.5 \\

    \midrule
    
    LLaVA-LLaMA-2-13B &40.1&38.8&27.8&26.5&8.9&8.3&27.8&26.7 \\

    \midrule
    Gemini Pro  &61.7&60.8 &42.5&41.9 &16.2&15.8 &43.2&42.5  \\
                
                \midrule
                
    Claude 3 Opus &57.3&56.7&37.8&37.2 &15.3&14.9&39.3 &38.7   \\
                
                \midrule
    Qwen-VL-Max & 62.5 &61.7 &47.2 &46.0 & 19.5 & 18.3 &46.5&45.4\\
            
\bottomrule
\end{tabular}}
\caption{Accuracy scores on our \ourbench{}. Element Accuracy is employed to gauge the comprehensiveness of elements, while Layout Accuracy is utilized to evaluate the effectiveness of webpage layout. These metrics are categorized into three difficulty levels: simple, medium, and complex.}
\label{maincomparison}
\end{table*}

\subsubsection{Performance by Complexity Level}
Through comparison, we gain the overall results. As complexity increases, the performance of all models decreases. Across all levels, WebSight emerges as the strongest model, consistently leading in both Element and Layout Accuracy. Gemini Pro and Qwen-VL-Max show good performance but with a greater drop as complexity increases. Most models like LLaVA-LLaMA-2-13B exhibit a more substantial decrease in performance as complexity increases, which may suggest that these models are better suited to simple tasks.

\subsubsection{Element vs. Layout Accuracy}
On average, across all complexity levels, element accuracy tends to be slightly higher than layout accuracy. This suggests that models are better at identifying and understanding individual elements than how those elements are arranged. The gap between two metrics tends to widen as the complexity of the task increases, which indicates that as tasks become more complex, it becomes more challenging for the models to understand of the layout information. WebSight is the top performer in balancing both element and layout accuracy.

\begin{table}[h]
\centering
\resizebox{0.95\linewidth}{!}{
\begin{tabular}{lcccccc}
\toprule
          &Ranking  & WebSight &Qwen-VL-Max &GP4V & Qwen-VL-Chat & miniGPT-4-LLaMA-2-7B \\
\midrule

        & Human  & 1  & 3 & 2  & 5 & 4 \\
\midrule
        & \ourbench{}  & 1  & 2 & 3 & 4 & 5\\
      
\bottomrule
\end{tabular}}
\caption{Ranking Comparison on Human Evaluation and \ourbench{}.}
\label{human evaluation}
\end{table}

\subsubsection{Impact of Enhancements}
Models with Five-hop MCoT all have an improvement than without Five-hop MCoT. It means Five-hop MCoT enhancement has a pronounced and positive impact on both element accuracy and layout accuracy across all levels of complexity, but the magnitude of the impact varies. The impact is most notable in the 'Simple' and 'Medium' complexity levels.

\subsubsection{Human-in-the-loop Evaluation}

First, we invite 20 front-end technology professionals to rate 100 web pages generated by different models. We meticulously design a questionnaire with multiple assessment dimensions. We compare the differences between the ranking based on our metrics and the ranking according to the average scores from the human evaluation panel. The Pearson coefficient between two rankings is 0.8 and P-Value is 0.104. Generally, when the coefficient is greater than 0.7, it is usually considered to be a strong correlation. The results are in Table~\ref{human evaluation}.

\section{Ablation}
In this section, we conduct the ablation on our Five-hop Multimodal Chain-of-Thought method, all experiments are conducted on GPT4V.

\subsection{Ablation on SoM prompt injection} 
\begin{table}[htp]
\centering
\resizebox{0.9\linewidth}{!}{
\begin{tabular}{lcccccccc}
\toprule
           Model &\multicolumn{2}{c}{Simple} &\multicolumn{2}{c}{Medium} &\multicolumn{2}{c}{Complex} &\multicolumn{2}{c}{Average} \\

           \cmidrule(lr){2-3}\cmidrule(lr){4-5}\cmidrule(lr){6-7}\cmidrule(lr){8-9}
           &EA &LA &EA &LA&EA &LA&EA &LA \\        
\midrule

        GPT4V w/ SoM  &69.4&66.4&43.1&42.6&16.7&15.9&45.8&44.5 \\
    \midrule
        GPT4V w/o SoM &64.5 &59.3 &38.6  &37.7 &14.2 &13.1 &41.6 &39.4 \\

\bottomrule
\end{tabular}}
\caption{Ablation on SoM module. We conduct experiments on GPT4V covering simple, medium, and complex levels. EA means element accuracy, LA means layout accuracy.}
\label{SoM}
\end{table}

In Table~\ref{SoM}, the results clearly demonstrate the significant impact of the SoM prompt injection module on the performance of GPT4V. When we conduct ablation by removing the SoM module from the architecture, we observe a substantial drop in performance. Specifically, GPT4V without the SoM module experiences a decrease of 4.2\% and 5.1\% on average element accuracy and layout accuracy, underscoring the crucial role that the SoM prompt injection module plays in enhancing the ability of models. As the complexity of the tasks increases, the accuracy of both models declines. The models perform better on simple and medium level.

\subsection{Ablation on Reflection}

\begin{figure}[ht]
    \centering {\includegraphics[width=0.69\linewidth]{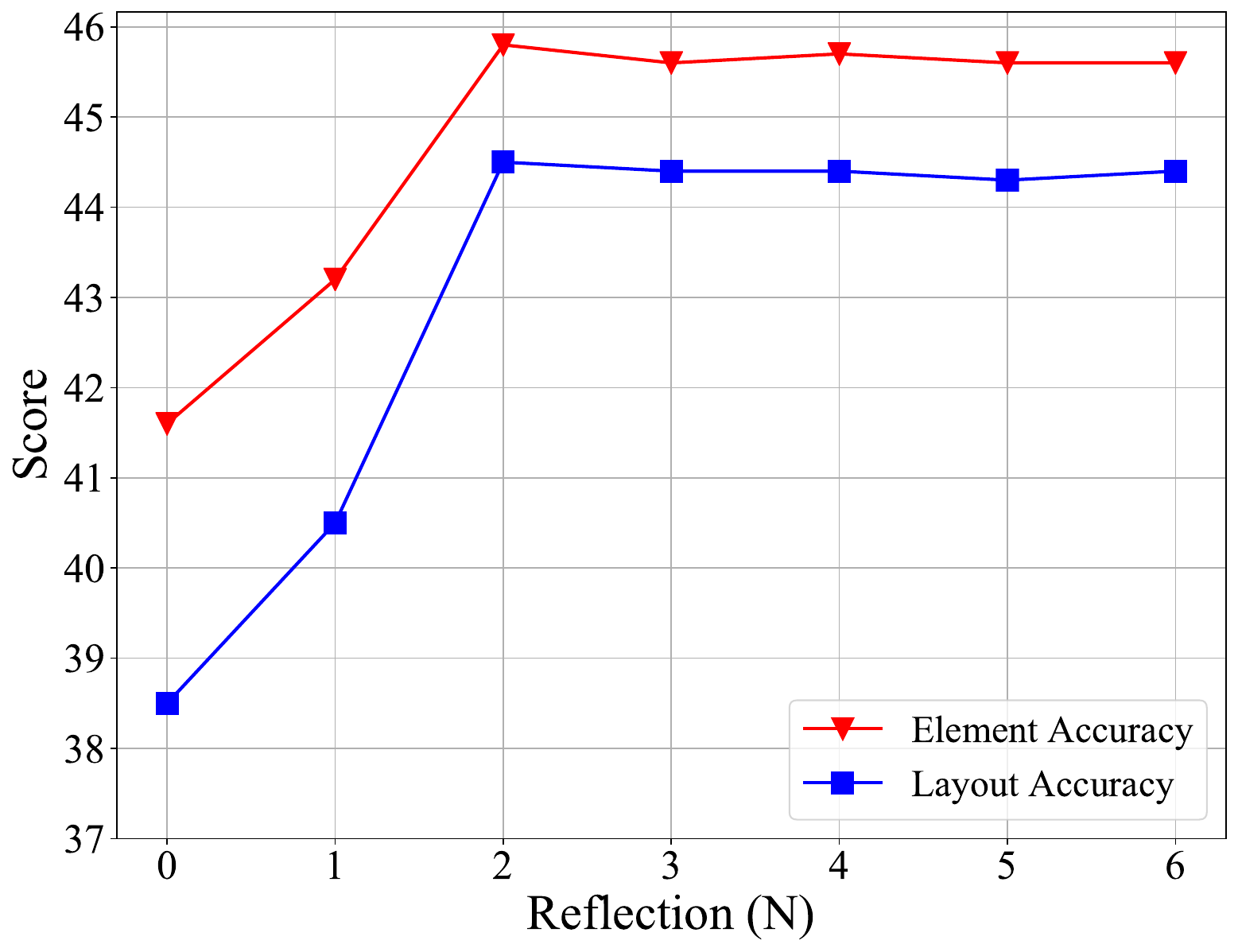}}
    \caption{Ablation on the times (N) of reflection. Element Accuracy and Layout Accuracy are calculated. We demonstrate the variation in scores as the number of reflections increases.}
    \label{fig:reflection}
\end{figure}

We conduct ablation experiments on the reflection module, and obtain the accuracy covering different times(N) of reflection in Figure~\ref{fig:reflection}. Element accuracy (red) shows a significant increase from N=0 to N=2, then stabilizes.
Layout accuracy (blue) exhibits minimal variation across the range of N values, with an overall stable trend. For most of the range, element accuracy are higher than those of layout accuracy. We can see that element accuracy and layout accuracy both improve significantly with an increase in Reflection iterations and then levels off. 

\begin{figure}[ht]
    \centering {\includegraphics[width=0.95\linewidth]{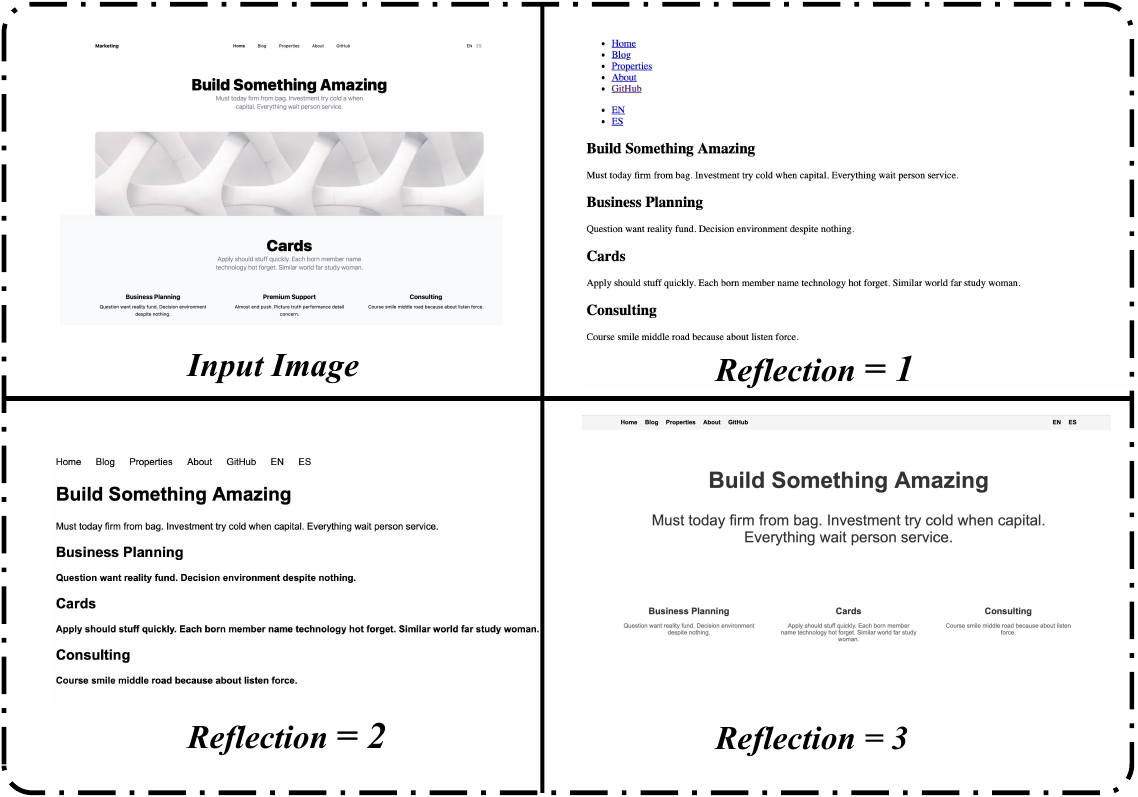}}
     \caption{Visualization on the reflection. The reflection times $N$=3. We can clearly see the element and layout improvements during different times of reflection.}
    \label{fig:visualization}
\end{figure}

\section{Visualization} 

In this section, we present web page renderings after various reflections, highlighting progressive enhancements. In Figure~\ref{fig:visualization}, the original design features a prominent title, three subheadings—Business Planning, Cards, Consulting—and the corresponding text. The navigation bar includes Home, Blog, Properties, About, GitHub, and language options (EN, ES). The first reflection accurately captures the structure of navigation bar, though the text does not completely match. The alignment of the main title and subheadings requires adjustment. By the second reflection, text fidelity improves, aligning closer with the original content, and the navigation bar layout becomes more compact. The third reflection brings further precision to text and tightens the overall page layout, aligning visual elements more accurately and reducing layout errors. The use of color and contrast also more closely mirrors the original, enhancing visual consistency.

\section{Conclusion}
In this paper, we introduce \ourbench{}. Two evaluation metrics are proposed: Element Accuracy to assess the completeness of elements, and Layout Accuracy to evaluate the positional relationships of elements. Furthermore, we outline a five-hop Multimodal Chain-of-Thought method aimed at enhancing image-to-web conversion. We evaluate large multimodal models and provide an analysis of the results.

\clearpage
\section*{Limitations}
\paragraph{Language Scope.} Currently, the benchmark is limited to only two languages. Expanding the scope to include additional languages would enhance its applicability and relevance to a more diverse global audience.

\paragraph{Data Quantity.} The benchmark dataset requires a significant increase in samples to ensure robustness and reliability. More comprehensive data coverage across different scenarios and contexts will improve the validity of benchmark results.

\bibliography{custom}

\appendix

\clearpage
\section{Data De-identification} \label{app:deid}
For real web pages, we perform de-identification to ensure sensitive information is not exposed:
\begin{itemize}
    \item The image components within our dataset have been replaced to avoid copyright or other legal concerns.
    \item For content that includes logos or is characteristic of certain brands, we have also made necessary modifications.
\end{itemize}
These measures ensure that identifiable elements are not directly recognized, thus adhering to copyright and trademark laws. This approach protects the interests of copyright holders while ensuring our project remains legal and ethical.

\section{Crowdsourcing}

In our study, we recognize several potential risks for participants. First, there is a risk to privacy and confidentiality since participants are required to share personal information. To address this, we anonymize all data and store it securely, ensuring access is limited to authorized personnel only. Second, psychological risks, such as discomfort or stress, may arise during the tasks. We mitigate these by providing clear instructions and holding debriefing sessions to support participants throughout the study. Additionally, participants are free to withdraw at any time without any penalties. While no significant physical risks are associated with our procedures, we continuously monitor for any signs of distress and offer appropriate support as needed. Each participant receives hourly compensation of \$20. Our primary participants for this study are web engineers.

\section{More Related Work \label{appendix:relatedwork}}
\subsection{Image-to-Web}
Recent advancements in the field of web development have seen researchers exploring innovative ways to convert webpage images into HTML code~\cite{imagecode1,imagecode3,imagecode7}. Researcher~\cite{imagecode1} focuses on automatically generating HTML code from webpage mock-ups. This involves the use of pix2code~\cite{pix2code}, which trains a CNN-based model on various webpage structure mock-ups, showcasing the potential of automating GUI creation. Another method~\cite{autoimagecode4} employs computer vision to identify objects and deep systematic analysis for result generation. Sketch2Code~\cite{imagecode6} divides the problem into three parts: object recognition, bounding box creation, and the creation of a functional prototype application. Recently, attention has shifted towards GPT4V for its image understanding and code generation capabilities. In the project screenshot-to-code~\footnote{https://github.com/abi/screenshot-to-code}, GPT4V~\cite{openai2023gpt4v} is used as the base model. WebSight~\cite{websight} from Huggingface is finetuned on numerous collected image-code pairs, however it has not conducted public performance evaluations. We desperately need a benchmark to evaluate large multimodal models in this domain.

\subsection{Multimodal Large Language Model}
Recent advancements in AI have led to the development of generative foundation models~\citep{bommasani2021opportunities} like GPT-3, ChatGPT, GPT-4, Claude and LLaMA~\citep{brown2020language,openai2022chatgpt,openai2023gpt4,claude2,touvron2023llama,llamaadapter2023}, which excel in a variety of text-based tasks without specific finetuning. Their performance has been evaluated across disciplines such as QA, math, and science~\citep{chen2021evaluating,sun2023scieval,wang2023scibench,huang2023c,huang2022language,liu2023agentbench}. 
On the vision-language side, there are several generative foundation models such as Qwen-Max, Qwen-VL, LLaVA, MiniGPT4, InstructBLIP, Flamingo,, Multimodal Bard~\citep{qwenvl,zhu2023minigpt,instructblip,alayrac2022flamingo,awadalla2023openflamingo,gao2023llamaadapterv2,google2023bard} that are trained on extensive image-text data, paving the way for multimodal learning~\citep{schuhmann2022laion,sharma2018conceptual,lin2014microsoft,zhu2023multimodal}. In addition, models specialized versions for document understanding are proposed~\citep{zhang2023llavar,ye2023mplug}. Benchmarks, like Visit-Bench, MMBench~\citep{bitton2023visit,yu2023mm,liu2023mmbench,xu2023lvlm,shao2023tiny}, have assess the instruction-following and reasoning capabilities. 
As these foundation models become more relevant to real-world applications, unlike prior work, we plan to benchmark their capabilities of the hot Image-to-Web area on a diverse set of visual contexts.

\subsection{Chain-of-Thought}
We have witnessed the remarkable capabilities of Large Language Models (LLMs), with their reasoning abilities significantly enhanced through approaches such as Chain-of-Thought (CoT)~\citep{wei2022chain}, Program-of-Thought (PoT)~\citep{chen2022program}, and Inductive Reasoning~\citep{wang2023hypothesis, tan2023large}. For multimodal CoT, recent work includes MCoT~\cite{mmcot}, HoT~\cite{hot}, CoCoT~\cite{cocot}. In this paper, we propose a five-hop multimodal Chain-of-Thought for evaluating multimodal large models on image-to-web domain while we compare other multimodal baselines.

\section{Prompt for Benchmark Generation\label{prompt:webgenerate}}

\subsection{General prompt for JavaScript Events}\label{promt:java}
\begin{mybox}
\textbf{Unified JavaScript Event}: 
onclick, onload, onmouseover, onmouseout, onchange, onsubmit, onmousemove, onmouseup, onmousedown, ondblclick, onkeydown, onkeyup, onkeypress, onsubmit, onfocus, onblur, oninput, onload, onresize, onscroll, onunload, ontouchstart, ontouchmove, ontouchend, onerror, oncontextmenu.
\end{mybox}

\section{List of Pre-defined JavaScript Event Bindings\label{app:jsevent}}
\begin{mybox}
    The total events are: 'onclick', 'onload', 'onmouseover', 'onmouseout', 'onchange', 'onsubmit'.
\end{mybox}

\subsection{General prompt for Domains\label{prompt:promptdomain}}

\begin{mybox}
\textbf{Domain Prompt 1}: Now that you are an HTML expert in designing websites, I will give you some requirements for designing a website. Please help design a static web page with a modern aesthetic for displaying a photographer's portfolio. Includes an image gallery with high-resolution pictures and smooth transitions, a detailed introduction page about the photographer's professional background and artistic philosophy, contact information page and quick links to the photographer's various social media accounts. The entire website should be responsively designed to adapt to the display of different devices; in addition, I will give you some other elements that need to be included in this web page. The elements to be included in brackets [] are [elements that need to be included]. In addition, some JavaScript (JS) functions need to be included. The JS functions to be included are in brackets [], that is, [JS functions to be included]. Please generate the corresponding website HTML code according to these requirements, and ensure that each website has CSS styles. It should be noted that the above text content must not be displayed directly in the generated web page to ensure that the HTML meets the requirements and is clean.
\end{mybox}

\begin{mybox}
\textbf{Domain Prompt 2}: Now that you are an HTML expert in designing websites, I will give you some requirements for designing a website. Please help me create a geo-targeted news website homepage that can display the latest local news, weather updates and emergencies based on the user's IP address. notify. The page design should be modern and user-friendly, including a dynamic news scroller, real-time weather widgets, and a personalized dashboard with user-customizable content. The page should also provide an advanced search function, allowing users to find news based on keywords, dates or categories; in addition, I will give you some other elements that need to be included in this page. The elements to be included are in brackets [], that is [ elements that need to be included]. In addition, some JavaScript (JS) functions need to be included. The JS functions to be included are in brackets [], that is, [JS functions to be included]. Please generate the corresponding website HTML code according to these requirements, and ensure that each website has CSS styles. It should be noted that the above text content must not be displayed directly in the generated web page to ensure that the HTML meets the requirements and is clean.

\end{mybox}

\begin{mybox}
\textbf{Domain Prompt 3}: Now that you are an HTML expert in designing websites, I will give you some corresponding requirements for designing a website. Please help create and design a portal for internal use of the enterprise, integrating key company resources and services. The website should include a dynamic press release section, a complete employee directory, internal forums to support communication and discussion among employees, workflow management tools, and a secure document sharing and collaboration platform. The website interface should be simple, easy to navigate, and have powerful search functions and personalization options; in addition, I will give you some other elements that need to be included in this web page. The elements to be included are in brackets [], that is, [need to include Elements]. In addition, some JavaScript (JS) functions need to be included. The JS functions to be included are in brackets [], that is, [JS functions to be included]. Please generate the corresponding website HTML code according to these requirements, and ensure that each website has CSS styles. It should be noted that the above text content must not be displayed directly in the generated web page to ensure that the HTML meets the requirements and is clean.

\end{mybox}

\begin{mybox}
\textbf{Domain Prompt 4}: Now that you are an HTML expert in designing websites, I will give you some corresponding requirements for designing a website. Please help build a web page for an online fashion clothing store that focuses on displaying the latest clothing trends. The webpage should contain multiple categories, such as 'New Products', 'Hot-Selling Items', and 'Discount Area'. Each product page should provide high-definition pictures, detailed product descriptions, size information, user reviews and a simple shopping process. The website should also support secure payment, order tracking and customer service chat functions; in addition, I will give you some other elements that need to be included in this web page. The elements to be included in brackets [] are [elements that need to be included]. In addition, some JavaScript (JS) functions need to be included. The JS functions to be included are in brackets [], that is, [JS functions to be included]. Please generate the corresponding website HTML code according to these requirements, and ensure that each website has CSS styles. It should be noted that the above text content must not be displayed directly in the generated web page to ensure that the HTML meets the requirements and is clean.

\end{mybox}

\begin{mybox}
\textbf{Domain Prompt 5}: Now that you are an HTML expert in designing websites, I will give you some corresponding requirements for designing a website. Please help design a blog website with the theme of personal travel and food experiences. Each blog should contain rich graphic content, such as detailed introductions to travel destinations, food recommendations, personal stories and travel tips. The website should include an interactive comments section that allows readers to leave comments and shares, and a section that showcases the best blogs of the month. The entire website should have optimized SEO functions and a design that adapts to different screen sizes; in addition, I will give you some other elements that need to be included in this web page. The elements to be included in brackets [] are [elements that need to be included] . In addition, some JavaScript (JS) functions need to be included. The JS functions to be included are in brackets [], that is, [JS functions to be included]. Please generate the corresponding website HTML code according to these requirements, and ensure that each website has CSS styles. It should be noted that the above text content must not be displayed directly in the generated web page to ensure that the HTML meets the requirements and is clean.

\end{mybox}

\begin{mybox}
\textbf{Domain Prompt 6}: Now that you are an HTML expert in designing websites, I will give you some corresponding requirements for designing a website and please help create an educational platform that offers a variety of online programming courses. Each course should have a detailed overview, learning objectives, video tutorials, downloadable practice materials, an online programming practice environment, and a forum where users can interact. The website should also include a personal achievement tracking system that allows users to see their learning progress and badges or certificates earned. In addition, a website directory needs to be constructed to focus on educational resources. The directory should have clear categories such as 'Online Courses', 'Academic Research', 'Learning Tools', and provide detailed descriptions and ratings for each link. The website should also include an efficient search function and user recommendation system; in addition, I will give you some other elements that need to be included in this web page. The elements to be included in brackets [] are [elements that need to be included]. In addition, some JavaScript (JS) functions need to be included. The JS functions to be included are in brackets [], that is, [JS functions to be included]. Please generate the corresponding website HTML code according to these requirements, and ensure that each website has CSS styles. It should be noted that the above text content must not be displayed directly in the generated web page to ensure that the HTML meets the requirements and is clean.

\end{mybox}
\begin{mybox}
\textbf{Domain Prompt 7}: Now that you are an HTML expert in designing websites, I will give you some corresponding requirements for designing a website. Please help design a social networking platform with a modern interface and powerful social functions. Users can create profiles, post status updates, share pictures and videos, and interact with friends. Platforms should offer advanced privacy settings that allow users to control who can see their content. In addition, there should be a recommendation system to display relevant content based on user interests and interactions; in addition, I will give you some other elements that need to be included in this web page. The elements to be included are in brackets [], that is, [need to include Elements]. In addition, some JavaScript (JS) functions need to be included. The JS functions to be included are in brackets [], that is, [JS functions to be included]. Please generate the corresponding website HTML code according to these requirements, and ensure that each website has CSS styles. It should be noted that the above text content must not be displayed directly in the generated web page to ensure that the HTML meets the requirements and is clean.

\end{mybox}

\begin{mybox}
\textbf{Domain Prompt 8}: Now that you are an HTML expert in designing websites, I will give you some corresponding requirements for designing a website. Please help build a website with the theme of technology news, providing the latest technology news reports, in-depth analysis articles, podcasts and video content. The page design should be modern and user-friendly, including a news scroller, video playback area, hot topics section, and a section for subscribing to the newsletter; in addition, I will give you some other elements that need to be included in this page. The elements to be included are within the brackets [], that is, [elements to be included]. In addition, some JavaScript (JS) functions need to be included. The JS functions to be included are in brackets [], that is, [JS functions to be included]. Please generate the corresponding website HTML code according to these requirements, and ensure that each website has CSS styles. It should be noted that the above text content must not be displayed directly in the generated web page to ensure that the HTML meets the requirements and is clean.

\end{mybox}

\begin{mybox}
\textbf{Domain Prompt 9}: Now that you are an HTML expert in designing websites, I will give you some corresponding requirements for designing a website. Please help design an environmentally-themed forum website that provides multiple different discussion sections, such as 'sustainable lifestyle', 'environmental protection Laws', 'Environmental Activities and Initiatives'. The forum should have a user-friendly interface that supports users to publish and edit posts, vote and participate in discussions supervised by moderators; in addition, I will give you some other elements that need to be included in this web page. The ones to be included are in brackets [] Element, that is, [the element that needs to be included]. In addition, some JavaScript (JS) functions need to be included. The JS functions to be included are in brackets [], that is, [JS functions to be included]. Please generate the corresponding website HTML code according to these requirements, and ensure that each website has CSS styles. It should be noted that the above text content must not be displayed directly in the generated web page to ensure that the HTML meets the requirements and is clean.

\end{mybox}

\begin{mybox}
\textbf{Domain Prompt 10}: Now that you are an HTML expert in designing websites, I will give you some requirements for designing a website. Please help me create an aggregation website that collects reviews of the latest movies and TV series. The website should include a live-updating comments section, user rating system, and direct links to viewing options on different streaming platforms. The page design should be concise and easy to navigate, allowing users to customize their content preferences; in addition, I will give you some other elements that need to be included in this web page. The elements to be included in brackets [] are [elements that need to be included] . In addition, some JavaScript (JS) functions need to be included. The JS functions to be included are in brackets [], that is, [JS functions to be included]. Please generate the corresponding website HTML code according to these requirements, and ensure that each website has CSS styles. It should be noted that the above text content must not be displayed directly in the generated web page to ensure that the HTML meets the requirements and is clean.

\end{mybox}

\subsection{Prompt for Simple-level Element generation}\label{prompt:simple}
\begin{mybox}
\textbf{Simple Prompt}: easy elements = ["title", "image", "icon", "card layout", "sliding banner/carousel", "footer", "sidebar", "background image and pattern"]
\end{mybox}

\subsection{Prompt for Medium-level Element generation}\label{prompt:medium}

\begin{mybox}
\textbf{Medium Prompt}: medium elements = ["Title", "Hyperlinks and Buttons", "Image", "Audio", "Sliding Banner/Carousel", "Card Layout", "Navigation Bar", "Footer", "Sidebar", " Breadcrumbs", "Background images and patterns", "Videos", "Social sharing buttons", "Progress bars and loading animations", "Comments area", "Tabs or accordions", "Modal windows/popups ", "Form", "Search bar"]
\end{mybox}

% \begin{figure}
%     \centering
%     \includegraphics[width=1\linewidth]{example_24.png}
%     \caption{Enter Caption}
%     \label{fig:enter-label}
% \end{figure}

\section{Prompt for Direct Web Generation\label{prompt:directgeneration}}

\begin{mybox}
    \textbf{Generation Prompt}: Generate the corresponding web code based on the image input
\end{mybox}

% Insert this content into an existing section of your LaTeX document

\section{Web Page Quality and User Experience Questionnaire}
We have designed this questionnaire to evaluate the quality of web pages and user experience across multiple dimensions. Please rate each statement on a scale from 1 to 5, where 1 represents 'very dissatisfied/very difficult to achieve' and 5 represents 'very satisfied/very easy to achieve.'

\begin{tcolorbox}[colback=gray!10, colframe=black!70, title=Questionnaire, fonttitle=\bfseries, boxrule=0.8mm, arc=2mm]

\begin{enumerate}[label=\arabic*.]
    \item \textbf{Content Comprehensibility}
    \begin{enumerate}[label*=\arabic*.]
        \item The content on the web page is easy to understand.
        \item The information provided is relevant and useful.
        \item The language used is clear and concise.
    \end{enumerate}

    \item \textbf{Layout and Structure}
    \begin{enumerate}[label*=\arabic*.]
        \item The layout of the web page is visually appealing.
        \item The structure of the web page is logical and easy to follow.
        \item The web page is well-organized.
    \end{enumerate}

    \item \textbf{Interactivity and Functionality}
    \begin{enumerate}[label*=\arabic*.]
        \item The interactive elements on the web page work as expected.
        \item The web page is responsive and loads quickly.
        \item The navigation is intuitive and user-friendly.
    \end{enumerate}

    \item \textbf{Overall Satisfaction}
    \begin{enumerate}[label*=\arabic*.]
        \item I am satisfied with my overall experience on the web page.
        \item I would recommend this web page to others.
        \item I am likely to return to this web page in the future.
    \end{enumerate}
\end{enumerate}

\textbf{Rating Scale}
\begin{itemize}
    \item 1 - Very dissatisfied/Very difficult to achieve
    \item 2 - Dissatisfied/Difficult to achieve
    \item 3 - Neutral
    \item 4 - Satisfied/Easy to achieve
    \item 5 - Very satisfied/Very easy to achieve
\end{itemize}

\textbf{Thank You}

Thank you for taking the time to complete this questionnaire. Your feedback is valuable and will help us improve the quality of our web pages and user experience.

\end{tcolorbox}

\section{Details of CSS Style\label{app:style}}
\begin{mybox}
\textbf{default values}: "none", "0", "normal", "0px",
                        "auto", "rgba(0, 0, 0, 0)", "rgb(0, 0, 0)"
                        
\textbf{keep properties}: 
        "color", "display", "font-family", "font-size", "height", "line-height", "margin-top", "text-align", "width",
        "background-color", "border-bottom-color", "border-bottom-left-radius", "border-bottom-right-radius",
        "border-bottom-style", "border-bottom-width", "border-image-outset", "border-image-repeat", "border-image-slice",
        "border-image-source", "border-image-width", "border-left-color", "border-left-style", "border-left-width",
        "border-right-color", "border-right-style", "border-right-width", "border-top-color", "border-top-left-radius",
        "border-top-right-radius", "border-top-style", "border-top-width", "box-shadow", "z-index",
        "margin-bottom", "margin-left", "margin-right", "padding-bottom", "padding-left", "padding-right", "padding-top",
        "position", "font-weight", "overflow-x", "overflow-y", "outline-color", "outline-style", "outline-width",
        "text-indent", "vertical-align", "background-attachment", "background-clip", "background-image", "background-origin",
        "background-position-x", "background-position-y", "background-repeat", "background-size", "border-style",
        "border-width", "box-sizing", "cursor", "font-feature-settings", "font-kerning", "font-optical-sizing",
        "font-variant-alternates", "font-variant-caps", "font-variant-east-asian", "font-variant-ligatures",
        "font-variant-numeric", "font-variant-position", "font-variation-settings", "letter-spacing", "opacity",
        "text-decoration", "text-decoration-color", "text-decoration-style", "text-emphasis-color", "text-emphasis-position",
        "text-overflow", "text-rendering", "text-shadow", "text-transform", "white-space-collapse", "word-spacing",
        "writing-mode", "align-items", "appearance", "background", "border", "flex-direction", "flex-shrink",
        "flex-wrap", "grid-auto-flow", "justify-content", "object-fit", "object-position", "overflow", "padding",
        "text-emphasis", "transform", "transition", "animation", "visibility", "white-space", "-webkit-font-smoothing",
        "-webkit-rtl-ordering", "-webkit-tap-highlight-color"
\end{mybox}

\section{More examples in \ourbench{}\label{case:all}}
\begin{figure}[h]
    \centering
    \includegraphics[width=0.8\linewidth]{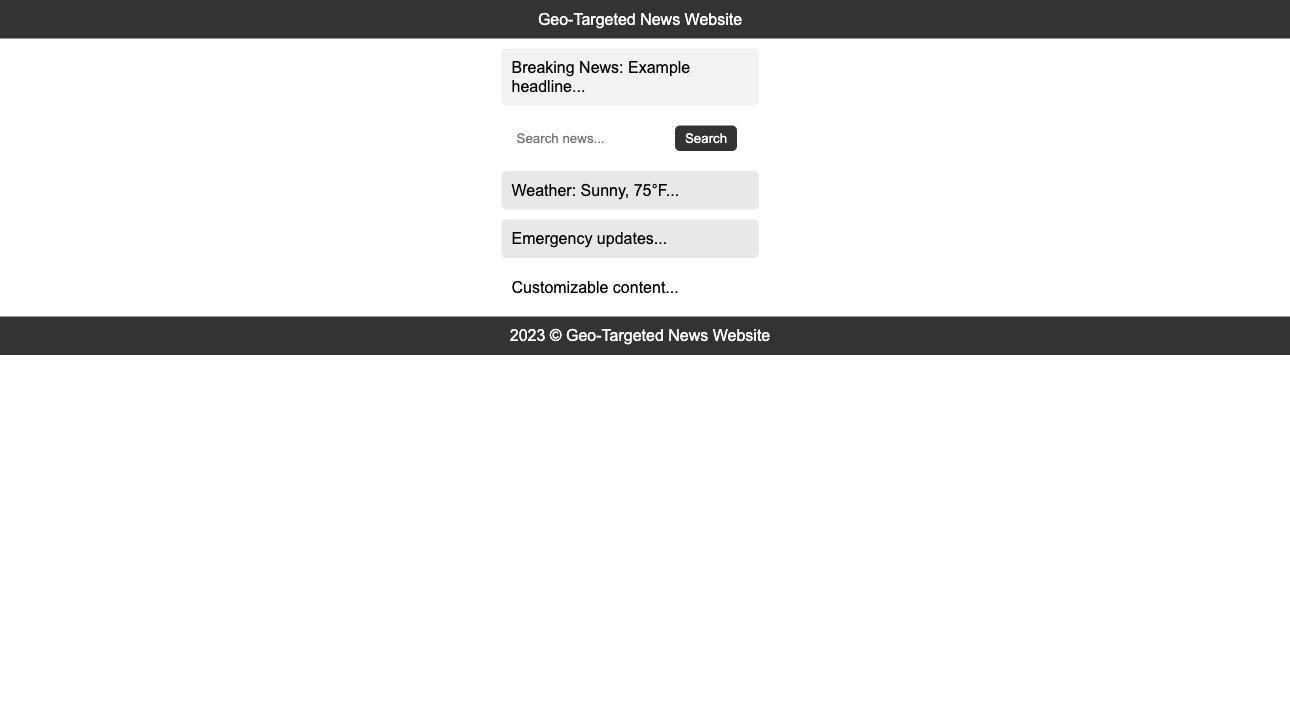}
    \caption{An Example of Simple Level}
    \label{fig:example_25}
\end{figure}

\begin{figure}[h]
    \centering
    \includegraphics[width=0.8\linewidth]{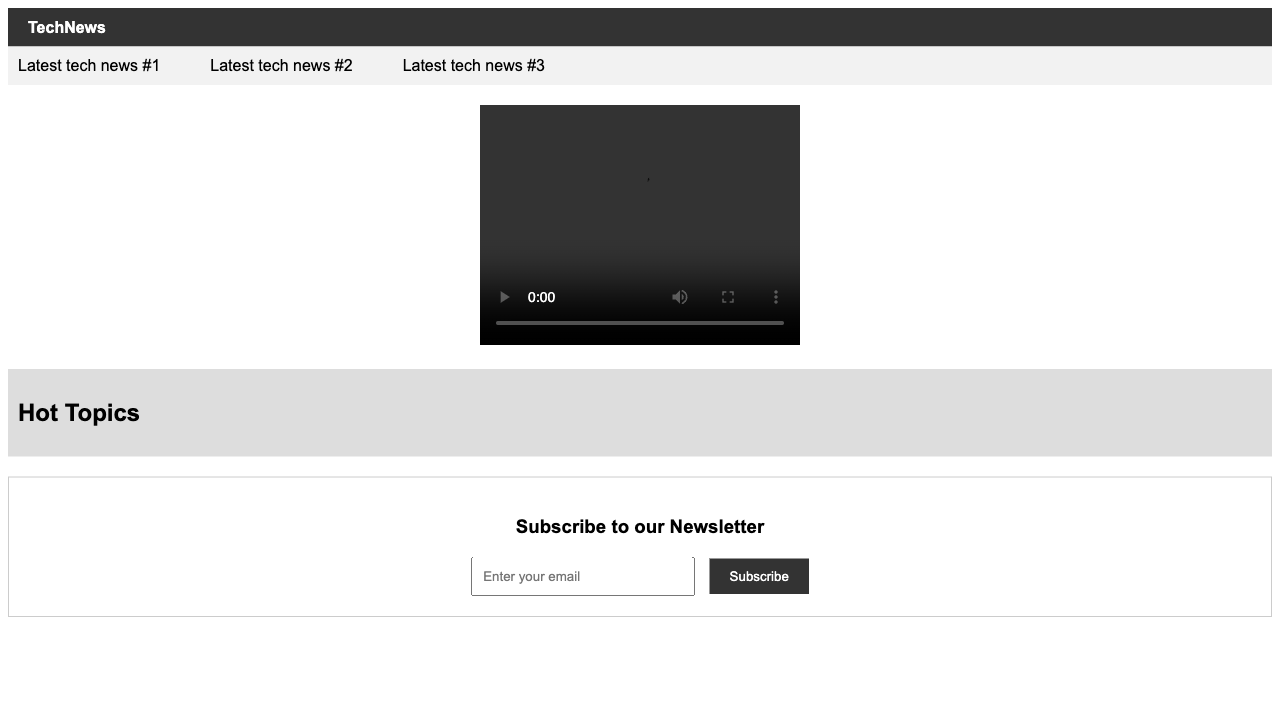}
    \caption{An Example of Medium Level}
    \label{fig:example_26}
\end{figure}

\begin{figure}[h]
    \centering
    \includegraphics[width=0.6\linewidth]{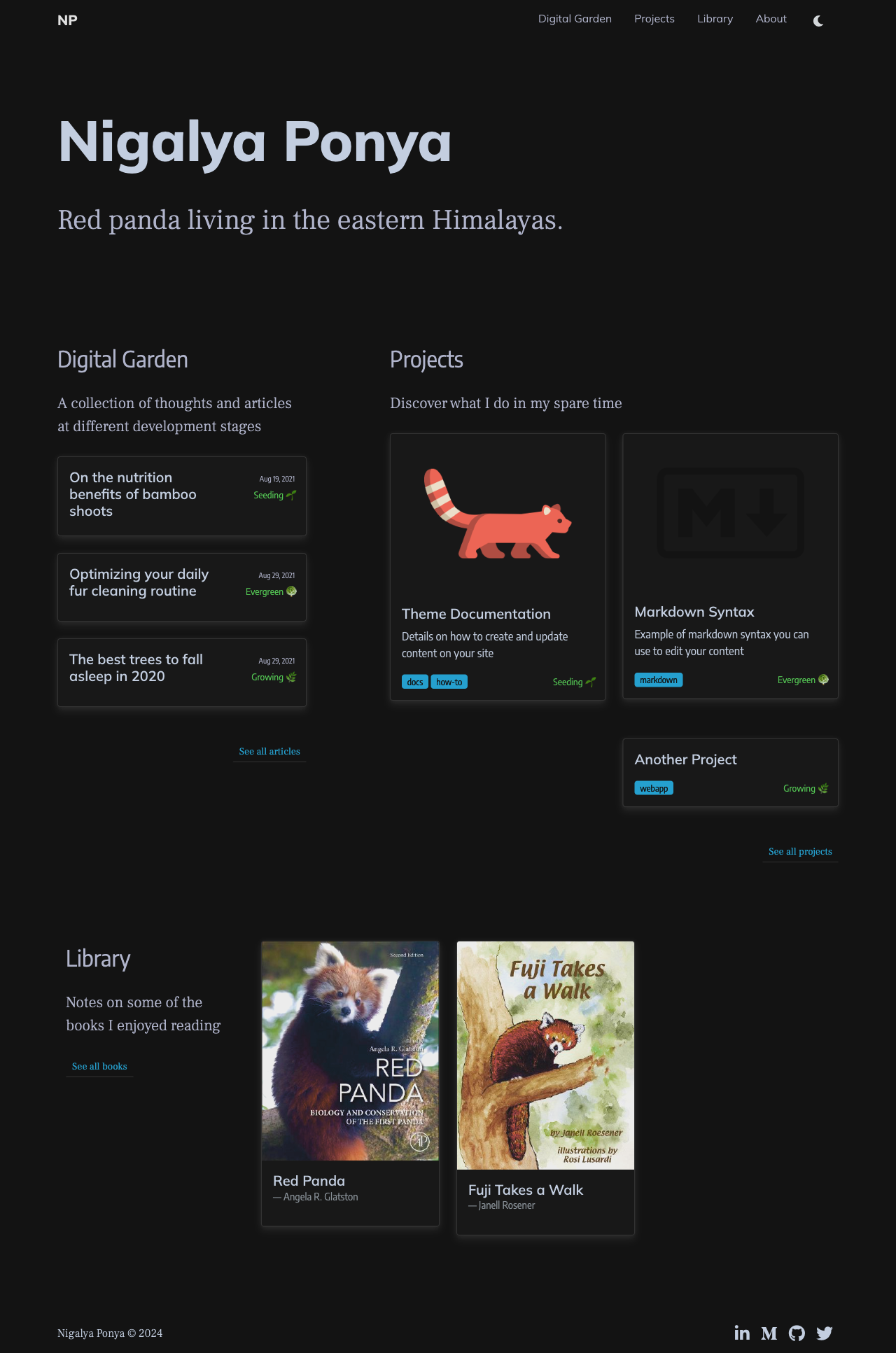}
    \caption{An Example of Complex Level}
    \label{fig:example_28}
\end{figure}

\section{List of Pre-defined Attributes of Elements\label{app:attributes}}

\begin{mybox}
\textbf{Attribute List1}:
        "a":{
            "text content": "strong comparison",
            "href": "weak comparison",
            "target": "weak comparison",
            "rel": "weak comparison",
            "download": "weak comparison",
            "hreflang": "weak comparison",
            "media": "weak comparison",
            "type": "weak comparison"
        },
        
        "img": {
            "alt": "strong comparison",
            "src": "weak comparison",
            "srcset": "weak comparison",
            "sizes": "weak comparison"
        },
        
        "button": {
            "text content": "strong comparison",
            "type": "weak comparison",
            "onclick": "weak comparison",
            "disabled": "weak comparison",
            "name": "weak comparison",
            "value": "weak comparison"
        },
        
        "input": {
            "value": "strong comparison",
            "placeholder": "strong comparison",
            "required": "strong comparison",
            "checked": "strong comparison",
            "readonly": "strong comparison",
            "type": "weak comparison",
            "name": "weak comparison",
            "min": "weak comparison",
            "max": "weak comparison",
            "step": "weak comparison",
            "pattern": "weak comparison"
        },
        
        "div": {
            "class": "weak comparison",
            "id": "weak comparison",
            "style": "weak comparison"
        },
        
        "h1": {
            "text content": "strong comparison",
            "class": "weak comparison",
            "id": "weak comparison",
            "style": "weak comparison"
        },
        
        "p": {
            "text content": "strong comparison",
            "class": "weak comparison",
            "id": "weak comparison",
            "style": "weak comparison"
        },
        
        "ul": {
            "class": "weak comparison",
            "id": "weak comparison",
            "style": "weak comparison"
        },
        
        "li": {
            "text content": "strong comparison",
            "class": "weak comparison",
            "id": "weak comparison",
            "style": "weak comparison"
        },
        
        "span": {
            "text content": "strong comparison",
            "class": "weak comparison",
            "id": "weak comparison",
            "style": "weak comparison"
        },
        
        "table": {
            "class": "weak comparison",
            "id": "weak comparison",
            "style": "weak comparison"
        },
        
        "thead": {
            "class": "weak comparison",
            "id": "weak comparison",
            "style": "weak comparison"
        },
        
        "tbody": {
            "class": "weak comparison",
            "id": "weak comparison",
            "style": "weak comparison"
        },
        
        "tr": {
            "class": "weak comparison",
            "id": "weak comparison",
            "style": "weak comparison"
        }
        
         "td": {
            "text content": "strong comparison",
            "class": "weak comparison",
            "id": "weak comparison",
            "style": "weak comparison",
            "colspan": "weak comparison",
            "rowspan": "weak comparison"
        },
        
        "th": {
            "text content": "strong comparison",
            "class": "weak comparison",
            "id": "weak comparison",
            "style": "weak comparison",
            "colspan": "weak comparison",
            "rowspan": "weak comparison",
            "scope": "weak comparison"
        }
         "label": {
            "text content": "strong comparison",
            "for": "strong comparison",
            "class": "weak comparison",
            "id": "weak comparison",
            "style": "weak comparison"
        },
        
        "select": {
            "name": "weak comparison",
            "required": "weak comparison",
            "multiple": "weak comparison",
            "class": "weak comparison",
            "id": "weak comparison",
            "style": "weak comparison"
        },
        
        "option": {
            "text content": "strong comparison",
            "value": "strong comparison",
            "selected": "strong comparison"
        }
\end{mybox}

\begin{mybox}
\textbf{Attribute List2}:
        "textarea": {
            "placeholder": "strong comparison",
            "required": "strong comparison",
            "readonly": "strong comparison",
            "name": "weak comparison",
            "rows": "weak comparison",
            "cols": "weak comparison",
            "class": "weak comparison",
            "id": "weak comparison",
            "style": "weak comparison"
        },
        "footer": {
            "class": "weak comparison",
            "id": "weak comparison",
            "style": "weak comparison"
        },
        "header": {
            "class": "weak comparison",
            "id": "weak comparison",
            "style": "weak comparison"
        },
        "article": {
            "class": "weak comparison",
            "id": "weak comparison",
            "style": "weak comparison"
        },
        "section": {
            "class": "weak comparison",
            "id": "weak comparison",
            "style": "weak comparison"
        },
        "nav": {
            "class": "weak comparison",
            "id": "weak comparison",
            "style": "weak comparison"
        },
        "aside": {
            "class": "weak comparison",
            "id": "weak comparison",
            "style": "weak comparison"
        },
        "figure": {
            "class": "weak comparison",
            "id": "weak comparison",
            "style": "weak comparison"
        },
        "figcaption": {
            "text content": "strong comparison",
            "class": "weak comparison",
            "id": "weak comparison",
            "style": "weak comparison"
        },
        "main": {
            "class": "weak comparison",
            "id": "weak comparison",
            "style": "weak comparison"
        },
        "hr": {
            "class": "weak comparison",
            "id": "weak comparison",
            "style": "weak comparison"
        },
        "br": {},
        "link": {
            "href": "weak comparison",
            "rel": "weak comparison",
            "media": "weak comparison",
            "type": "weak comparison"
        },
        "meta": {
            "content": "strong comparison",
            "name": "weak comparison",
            "http-equiv": "weak comparison",
            "charset": "weak comparison"
        },
        "script": {
            "src": "weak comparison",
            "type": "weak comparison",
            "async": "weak comparison",
            "defer": "weak comparison"
        }
\end{mybox}

\end{document}